\documentclass[]{article}
\usepackage{arxiv}
\usepackage[T1]{fontenc}
\usepackage{graphicx}%
\usepackage{multirow}%
\usepackage{amsmath,amssymb,amsfonts}%
\usepackage{amsthm}%
\usepackage{mathrsfs}%
\usepackage[title]{appendix}%
\usepackage{xcolor}%
\usepackage{textcomp}%
\usepackage{manyfoot}%
\usepackage{booktabs}%
\usepackage{algorithm}%w
\usepackage{algorithmicx}%
\usepackage{algpseudocode}%
\usepackage{listings}%
\usepackage{todonotes}
\usepackage{csquotes}
\usepackage{lmodern}
\usepackage{natbib}

\setcitestyle{authoryear, open={(},close={)}}
\date{}
\MakeOuterQuote{"}
\usepackage[hyphens]{url}
\usepackage{hyperref}
\usepackage[hyphenbreaks]{breakurl}
\begin{document}

\title{Evaluating AI Evaluation: Perils and Prospects}


\author{John Burden\\ Leverhulme Centre for the Future of Intelligence, Centre for the Study of Existential Risk,\\ University of Cambridge,\\ jjb205@cam.ac.uk}




\maketitle

\begin{abstract}
    As AI systems \textit{appear} to exhibit  ever-increasing capability and generality, assessing their true potential and safety becomes paramount. This paper contends that the prevalent evaluation methods for these systems are fundamentally inadequate, heightening the risks and potential hazards associated with AI. I argue that a reformation is required in the way we evaluate AI systems and that we should look towards cognitive sciences for inspiration in our approaches, which have a longstanding tradition of assessing general intelligence across diverse species. We will identify some of the difficulties that need to be overcome when applying cognitively-inspired approaches to general-purpose AI systems and also analyse the emerging area of "Evals".  The paper concludes by identifying promising research pathways that could refine AI evaluation, advancing it towards a rigorous scientific domain that contributes to the development of safe AI systems.
\end{abstract}


\section{Introduction}
Recent years have seen an explosion of interest in mitigating the risks from AI systems, both existential and socio-technical \citep{ hendrycks_overview_2023, amodei_concrete_2016, critch_ai_2020, bostrom_superintelligence_2014,  dobbe_toward_2024, huang_overview_2023}. 
But how can we \textit{actually} ensure that an AI system is safe? This is a difficult and multi-faceted question, requiring conscious intervention at every step in the process of creating an AI system, from data-set collection, all the way through to deployment (and beyond).  One important part of this process is to develop techniques for 
\textit{building} and \textit{training} safe AI systems, or \textit{aligning} systems with particular sets of values \citep{russell_human_2019, gabriel_artificial_2020}. However, equally important is identifying whether these techniques have been successful, and subsequently deciding whether a system is ready for deployment. This is the domain of AI evaluation. 

Improved methodologies for both aligning and evaluating AI  are necessary to ensure that the seemingly ever-advancing AI systems are safe, ethical, and ultimately beneficial for humanity. This necessity is driven by the increasing capability and generality of AI systems. The success of Large Language Models (or Foundation Models \citep{bommasani_opportunities_2022}) such as the GPT series \citep{brown_language_2020,openai_gpt-4_2023} has demonstrated that a single (albeit incredibly large) model can be trained to perform a wide variety of tasks. These models can be "fine-tuned" at low-cost (relative to training a new model) to improve performance on specific tasks. They also demonstrate the ability to adapt to new tasks with just a few examples through a process called "few-shot learning" \citep{brown_language_2020}.

An increase in generality adds more than just a fixed, finite number of new use cases. Tasks can be combined or specified in arbitrary ways, for example solving a mathematics equation through poetry in Latin. The task-space over which we need to evaluate the system expands from almost a single point to a large and multi-dimensional vector-space. From a safety perspective, or even just to guarantee a minimum level of performance, this is orders of magnitude more difficult: we cannot simply evaluate performance on every individual task combination. We can't know a priori whether a new point in task-space will yield an unsafe response.

Complicating matters further, tasks can be seen as composed of multiple task \textit{instances}. For example, a task could involve finding paths through a maze; there are infinite variations of mazes with different difficulties, each yielding different responses from the same system. Consider evaluating the chess-playing system, \textit{Deep Blue} \citep{campbell_deep_2002}. What would it mean for a hypothetical deployment if the system could defeat a world champion but be defeated by an amateur? Is this system "good at chess"? If there are clusters of instances (individual games against specific opponents) where the system performs poorly, how do we reconcile that with any notion of competency? Particularly if these performance patterns violate our understanding of difficulty (i.e., we might expect the system to perform better against chess players with low Elo rankings).

Although evaluations of seemingly simple tasks are more complex than initially anticipated—task-space for maze solving or chess are already vast multidimensional vector-spaces---we see that more open-ended and general task-spaces will expand even further. As the number of dimensions of task-space increases, the number of evaluation samples needed rises exponentially with the number of dimensions to attain a fixed level of evaluation coverage. This is \citet{bellman_dynamic_2010}'s \textit{Curse of Dimensionality}. This \textit{Curse} is well-known and acknowledged in training ML systems, but it affects us twice because evaluation is subject to the same growing sample requirements.

Robust evaluation alone is not sufficient to ensure AI systems are safe. However, as I will argue, it is \textit{necessary}. Evaluation cuts across a wide swathe of areas within the development of safe AI, and improvements in evaluative processes can greatly propagate benefits. To make these improvements, we need to know what we want the future of AI evaluation to look like and what we want to avoid. We need an idea of both the perils and prospects. This is particularly important for dealing with more advanced AI systems, such as so-called AGI, not just in terms of the impacts these systems may have, but also in terms of how we evaluate systems of increasing intelligence.

In this paper, I argue that the current state of AI evaluation is dangerously flawed—our current trajectory is towards peril. I further want to highlight certain evaluation methodologies from the cognitive sciences, from which AI evaluation can greatly benefit, as well as the limitations of these techniques when directly applied to AI systems that need to be overcome. These are our "prospects". I end the paper by highlighting what I perceive to be the most promising research avenues to better improve evaluation techniques.

\section{Formalising Tasks, Instances, and Performance}

Formalising tasks, instances, and performance will allow us to more precisely discuss the problems and solutions in evaluating general-purpose AI systems and their impact on safety. Following \citet{hernandez-orallo_measure_2017}, we consider a task $M$ comprised of a set of instances $\mu$. The variation in instances can be parameterised, and this parameterisation constitutes the task-space of $M$. As a shorthand, we can write this as $\mu = M(x_1,..., x_n)$ for some set of parameters defining the variation in $M$. We assume that the task-space is deterministic and that $M$ forms a bijective function. That is, given parameters $[x_1, ..., x_n]$, there is always and only one instance $\mu = M(x_1, .., x_n)$. If both $\mu_1$ and $\mu_2 = M(x_1, .., x_n)$, then either $\mu_1 = \mu_2$ or our task characterisation is incomplete in its parameters.\footnote{This simplifying assumption is more for convenience of notation rather than making any real assumption about tasks or instances. That is, we want to be able to denote task instances individually.}

Revisiting our maze example, we can imagine $M$ taking parameters corresponding to the width and height of the maze, as well as parameters determining the layout of the walls and the location of the "goal".

Separate from, but often highly related to, the task $M$ is the distribution $p_M$ from which $\mu$ is sampled. The distribution $p_M$ may represent a natural preponderance of particular instances, or an intentional intervention to focus efforts on more important instances. Considering the maze example again, certain variations of mazes may be more likely than others; for example, we may prefer mazes that are solvable or within a certain range of sizes.

A system $\pi$ will produce a (possibly stochastic) response on instance $\mu$, denoted as $R(\pi, \mu)$, which is a numerical value. We do not limit $R$ to producing a scalar response; $R$ can return a vector of responses, such as both a performance metric and a safety metric. Given that the response $R(\pi, \mu)$ is often stochastic, we will frequently refer to the expected value of this response, denoted by $\psi(\pi, \mu) = \mathbb{E}[R(\pi, \mu)]$. This $\psi$ is often called the \textit{performance} of $\pi$ on $\mu$. Returning to our recurring maze example, we may obtain a collection of responses from a system $\pi$ interacting with $\mu$. The response could be related to the success or failure of solving the maze within a specified time limit (returning a collection of $0$s and $1$s). The expected value $\psi(\pi, \mu)$ would then correspond to the average rate of success of the agent over this collection of trials.

This relatively simple formulation encapsulates evaluation over a wide range of AI paradigms and evaluation criteria. The interesting and challenging aspects of AI evaluation arise not from these raw responses, but from how we interpret, aggregate, and use them, and how response patterns and potential system deployment fit into the environment in which they are placed. We will examine typical practise for evaluation of AI systems in Section \ref{sec:EvalAI}.

\subsection{Capability-oriented Evaluation and Performance-oriented Evaluation}

I will briefly distinguish between two evaluation styles: performance-oriented and capability-oriented \citep{burnell_not_2022, burden_inferring_2023}\footnote{Capability-oriented evaluation is sometimes referred to as feature-oriented \citep[Chapter 5, pg. 146]{hernandez-orallo_measure_2017}, ability-oriented \citep{hernandez-orallo_evaluation_2017}, or construct-oriented \citep{wang_evaluating_2023}}. Broadly speaking, performance-oriented evaluation assesses how well a system performs on a particular test, while capability-oriented evaluation measures the latent factors of the system (operationalised as capabilities) that cause differences in test performance. This view of capabilities broadly corresponds to the Conditional Analysis of Model Capabilities \citep{harding_what_2024}, where capabilities are attributed to a model if the model succeeds at $x$ when its output is best described as being directed to do $x$.

System capabilities are often difficult to directly measure---especially compared to directly observable properties like physical size. The presence or value of these latent capabilities must be inferred. Tests must be designed to discriminate between test subjects based on this latent capability. A key aspect of capability-oriented evaluation is the relationship between the capability being measured and the demands placed on that capability by a task instance. In other words, the difficulty of the task instance is crucial. This is at odds with a performance-oriented evaluation approach, which focuses on raw performance on the test itself.

Let's revisit our maze example to highlight the differences between these two evaluation paradigms. Consider a test battery of mazes comprised of multiple different task instances. Performance-oriented evaluation would report a metric related to the tests themselves, such as the percentage of occurrences a single agent achieved success over the test set. Capability-oriented evaluation, on the other hand, tries to report results related to inherent properties of the agent (or consistent features of the task instances that the agent can handle). With our continuing maze example, capability-oriented evaluation would report something like a "maze solving ability". This could be broken down further into more specific factors such as "obstacle handling" and "efficient exploration ability" or similar. Capability-oriented evaluation doesn't need to rely on high-level abilities: we could also see that the system has a "size" capability, relaying the maximum size of the maze that the system could reliably solve. We can easily conceive of further intrinsic features from the tasks that we can use to characterise the boundaries of the system's capabilities.

Capability-oriented evaluation has several key advantages over the performance-oriented alternative. For instance, what does it mean if a system achieves a particular success rate on a test? Without knowledge of the demands placed on the system by the test's constituent instances the result \textit{by itself} is often of little value. Performance-oriented evaluations can provide value in comparing test-takers --- it seems likely that a system receiving a score of 80\% is better than a system that only manages 50\%, but this doesn't inform us how capable a system may be in itself. The test may have been comprised of trivially easy instances, and neither system is actually very good, or the opposite could be true! Comparing two systems with a performance-oriented evaluation may also be limited in utility; two test-takers that achieve 99\% may not have the same capabilities if the test isn't able to discriminate well between high-performers.  

 On the other hand, under capability-oriented evaluation, the reported capability shouldn't change if the distribution of instance difficulty changes. Clearly, capability-oriented evaluation has the potential to be more informative about the system being assessed; however it is generally more difficult to perform such an evaluation because it requires a strong understanding of the relationship between the capability, instance difficulty, and how these affect observable performance is required. This is because many capabilities that we wish to assess about AI systems (and intelligent systems in general) are not directly observable. Evaluators can only observe performance on specific tasks that require certain elements of the capability in question (entangled with many others). Therefore, in order to properly assess a given capability, it needs to be inferred from a variety of performance data. To do this effectively requires a detailed understanding of the capability, as well as how it affects performance on a variety of task-instances.

 Despite this additional difficulty, throughout this paper I want to emphasise the benefits (and what I ultimately argue to be the necessity) of capability-oriented evaluation for AI---particularly from a safety perspective.

\subsection{The Fallacy of Reification?}
The capabilities we aim to assess using capability-oriented evaluation are, as discussed, not directly measurable. Capabilities are often abstract ideas such as "object permanence", "navigational skills", "language understanding", and so on. The "fallacy of reification" (sometimes called the "fallacy of misplaced concreteness") refers to treating abstract entities and ideas as if they were real, concrete entities \citep{whitehead_science_1925}. Debate about whether abstract objects "exist" has existed in philosophy for a long time (see, e.g., \citep{falguera_abstract_2022} for a summary of abstract objects and their contentious history). However, the "fallacy of reification" does not argue about whether abstract concepts \textit{exist}, but rather claims that treating an abstract concept (whether it exists in any meaningful way or not) as a concrete object is fallacious.

However, an important tool in science is the \textit{Hypothetical construct}. These are explanatory variables or factors that are not themselves observable \citep{maccorquodale_distinction_1948}. These \textit{constructs} are ubiquitous (gravity, motivation, intelligence are all constructs) within science. In some sense, these constructs are forms of reification. We say that objects fall to the ground because they are "pulled down by gravity"\footnote{A more "scientific" way of phrasing this would involve reference to a gravitational field, but these too, are constructs.}. We identify a \textit{physical location} on an object as its centre of mass. 
In short, we treat constructs as concrete and reify them. Obviously, some concepts (such as gravity or centre of mass) are more validly reifiable than others (e.g., the way we personify nature "Mother Nature abhors a vacuum"). How do we know when our hypothetical constructs are fallacious, and when they are justified? 
This primarily depends on two factors that together can enable us to identify constructs and reifications that are useful and representative of meaningful, \textit{real} phenomena. The first is a shared understanding of the construct's meaning as argued by \citet{zeigler-hill_hypothetical_2020}. The second is the validity of the construct. Construct validity is the extent to which the constructs actually measure what they claim to measure \citep{cronbach_construct_1955}. Also of importance is the notion of construct legitimacy: the extent to which the theory arguing for the construct is justified \citep{stone_defense_2019}. Today, construct validity is seen as an overarching term for many types of validation approaches \citep{messick_validity_1994}. One type that is particularly worth highlighting is what \citet{cronbach_construct_1955} refer to as \textit{predictive validity}: how well does this construct predict future scores on a particular test? Within AI evaluation and safety, prediction is paramount (I argue this more fully in section \ref{sec:prediction})  and predictive validity should be at the forefront of any evaluator's mind. 

Another crucial type of validity is \textit{external validity} \citep{campbell_experimental_1963}. External validity broadly corresponds to the extent to which the conclusions of a study can be generalised outside the context in which the study took place. With constructs, we need to be mindful not only of the construct validity of tests, but also of their external validity for measuring that construct. That is, even if the test validly measures the construct in one scenario, does it generalise to others? Different populations of subjects, environmental factors, and (sometimes seemingly minor) experimental details can drastically affect a test's ability to validly measure a construct. As an intuitive example, if we imagine a hypothetical test that has high construct validity for measuring intelligence that has been rigorously tested on humans, but then we give the test to a dog, we would likely find that the dog would not score any points on the test, despite dogs clearly having some level of intelligence. The dog was unable to complete the test, at least in part, because it cannot read or write. Even though the test has high construct validity \textit{when applied to humans}, the test lacks external validity outside the domain in which it was designed.\footnote{This example makes a subtle assumption that "intelligence" in humans is the same as "intelligence" in dogs. \citet{burkart_evolution_2017} argue there is evidence for general intelligence "g" explaining intra-species variation as well as a "G" factor explaining inter-species variation. On the other hand, \citet{poirier_how_2020} argue the evidence for "g" in non-human animal populations is weak. Regardless, the importance of external validity should be clear.} In \citet{messick_validity_1994}'s categorisation, external validity is encapsulated by his notion of construct validity. In this manuscript, for clarity, I will refer to external validity explicitly when it is the property I am referring to, but I follow Messick in that when I refer to construct validity I am implicitly requiring there to be external validity to the population being tested.

When evaluating AI systems, two forms of external validity are particularly relevant. The first relates to external validity across subjects: are pre-existing tests valid when applied to AI systems? Or, like the dog taking the intelligence test, is there a mismatch between the subject and the test that invalidates our results? The second form of external validity is situational: will the test inform us how well a system will perform when deployed "in the real world"? Or is our evaluation limited to artificial laboratory conditions?

Construct validity is crucial for capability-oriented evaluation because the indirectly measurable capabilities we are interested in are inherently hypothetical constructs that must be reified. Throughout this paper, I use the term "reify" to refer to the creation of a construct. This is partly an effort to "de-stigmatise" the term: not all constructs or reifications are fallacious; they simply need validation. This term also serves as a reminder of the opposite: we need to be careful not to confuse the map for the territory, remaining keenly aware of the need for legitimacy and validation.

\subsection{Evaluation Is For Prediction}\label{sec:prediction}
I argue that a core aspect of evaluation is \textit{prediction}. 
First, why do we evaluate systems (AI or otherwise)? The immediate answer is "to determine if the system is suitable for its purpose". While true, this overlooks the fact that we expect our evaluations to provide insight into how the system will perform outside of the evaluation, during "deployment".

When we decide that a system is "fit for purpose", we are \textit{predicting} that it will perform at an acceptable level in future instances of the task. This can be a supervisory process ("Is this system good enough for the task it was designed for?") or a reflective process ("What could be done differently in the future to improve performance on similar tasks?"), but in either case we are concerned with anticipating future behaviour and performance. We rely on this implicitly, often without realising it, in assessment scenarios. For example, we utilise standardised testing for university applications because we believe these tests demonstrate that a candidate has subject knowledge, work ethic, general problem-solving skills, and so on. But more importantly, we expect these factors to indicate—to predict—future success at the university.

The same goes for AI. We want to ensure that the AI systems we create are capable and safe. We perform various evaluations and derive performance metrics—even the less informative ones such as mean performance—because we believe these metrics capture important properties of the system that predict whether it will be fit for purpose: capable, reliable, and safe. This belief needs to be more fully and explicitly expressed in our evaluation instruments.

\section{Risks From Poor Evaluation}
What are the risks of poor evaluation methodology for AI? First, let's examine the behaviour we are trying to avoid. Many undesirable characteristics have been identified as areas of concern for AI systems: negative side-effects \citep{amodei_concrete_2016}, power-seeking tendencies \citep{carlsmith_is_2022}, mesa-optimisation \citep{hubinger_risks_2021}, exacerbating bias \citep{ntoutsi_bias_2020}, and more. These characteristics are often framed as natural occurrences of optimisation (e.g., power-seeking behaviour naturally arises under certain conditions in Markov Decision Processes \citep{turner_optimal_2021}) or reward mis-specification \citep{krakovna_specification_2018}. Conscious interventions must be taken in the training and data-curation process to disincentivise these negative characteristics. However, these interventions also need evaluating; the claims they make must be verified, and any trade-offs with other characteristics must be identified. 

Flawed evaluations may lead to undue confidence in strategies for system alignment or addressing safety issues. This could result in unsuitable deployment in safety-critical domains and cause harm. This type of risk also includes overconfidence in the \textit{absence} of certain characteristics (e.g., bias or deceptive behaviour). These are two sides of the same coin, often framed as separate issues. Concerns about whether a system is robust and reliably safe are arguably the same as concerns about the potential for unsafe behaviour. Evaluations of both issues should be focused on reducing uncertainty about the presence of particular system characteristics related to the consistency of behaviour.

Consider a system that was trained to complete a task and was subsequently deployed. However, after deployment, the system begins to act in unexpected ways, such as failing to complete the task or exhibiting undesirable characteristics. How could this have arisen? This could have occurred due to a failure in the evaluation methodology (perhaps too little evaluation and testing was done or statistical techniques were misapplied). Alternatively, this could have occurred as an Out-of-Distribution (OoD) error, where the task distribution $p_M$ during testing/deployment differs from $p_M$ during training. Regardless of \textit{why} the system began acting in an unexpected manner, the mere fact that such a system was deployed represents a failure of the evaluation process. Not only was the system not "fit for purpose", but its behaviour wasn't predictable in the deployment environment.

As AI systems become both more capable and general, they will likely be deployed in more safety-critical domains. Governments are already pushing for this, such as the UK government advocating for a "pro-innovation" approach to AI and expressing a desire to see AI used in healthcare, policing, and transport \citep{uk_government_pro-innovation_2023}. Simultaneously, the increase in generality makes encountering out-of-distribution instances much more likely—systems such as GPT-4 are being used for an extremely diverse variety of tasks \citep[e.g., see][]{bubeck_sparks_2023}. This significantly increases the likelihood of encountering task instances from outside the training distribution. Hence, the problem of ensuring systems are robust to OoD errors is becoming increasingly important.

Robust and accurate evaluation techniques are essential to mitigate the risks originating from, for example, OoD errors. These techniques help us understand and predict how AI systems will behave in real-world environments, which often present scenarios not encountered during training. Without effective evaluation methods, we risk deploying AI systems that may behave unpredictably or unsafely when faced with novel situations, undermining trust and potentially causing harm. Improving our evaluation methodologies is a critical step towards ensuring the safe and reliable deployment of AI systems in society.

\section{Evaluation of AI systems in Practice}\label{sec:EvalAI}
In previous sections, I've discussed high-level ideals for AI evaluation, highlighting the importance of construct validity and predictability, and extolling the benefits of capability-oriented approaches. Now we will explore the stark contrast that is AI evaluation praxis. I will argue that the majority of these techniques are \textit{performance-oriented} evaluation. 
Traditional AI evaluation techniques often encapsulate performance as the expected response: 
\[ \Psi(\pi, M) = \mathop{\mathbb{E}}_{\mu \sim p_M} [\psi(\pi, \mu)] =  \int_{\mu \in M} p_M(\mu)\psi(\pi, \mu) d\mu \]
This is the mean response with respect to some task distribution. Recall that $\psi$ is the expected value of response $R$ of system $\pi$ on instance $\mu$. Given that we don't simply have $\psi$, $p_M$, or $R$  readily available, we need to work with sample estimates of this expected response.  
We receive sample of responses $\hat{R}$, and settle for the sample mean of a test distribution or benchmark:
\[ \hat{\Psi}(\pi, M) = \frac{1}{N}\sum_{i=1}^N\hat{\psi}(\pi, \mu_i) \]

Where $[ \mu_i \mid i \in \{1..., N\} ]$ is the list of test instances.  
Here $\hat{\psi}$ is the sample mean of the observed response $R$ of $\pi$ on $\mu_i$. Ideally to get a clearer picture of $\hat{\psi}$ we need to see $\pi$'s response on $\mu$, $R(\pi,\mu)$, multiple times to get an accurate estimate for $\hat{\psi}$

This simple approach captures the evaluation of a wide range of AI systems, from image classifiers ($R(\pi, \mu) = 1$ if $\pi$ classifies $\mu$ correctly, and $0$ otherwise, where $p_M$ is the distribution of test images) to reinforcement learning ($R(\pi, \mu)$ yields the agent's return on instance $\mu$, and $p_M$ is the distribution of test environments).

Often, other domain-relevant metrics are utilised as well, such as the F1 measure for balancing precision and recall \citep{chinchor_muc-4_1992, van_rijsbergen_information_1979}, or the BLEU score \citep{papineni_bleu_2002}. These too can be defined in terms of appropriate response functions. All of these metrics \textit{aggregate} performance results and eviscerate any information that may be used by the evaluator to better understand the system or predict responses on new instances.

Limited forms of capability-oriented evaluation have been explored in what is now AI's deep history. These included the Newell test \citep{anderson_newell_2003} and the Cognitive Decathlon \citep{mueller_bica_2007, mueller_adapting_2008, simpson_refining_2008}. However, these efforts often focused on identifying capabilities required for AI systems to solve problems and designing test suites to target these capabilities, rather than identifying methods to directly measure the relation between capabilities and task instance performance. The capabilities identified weren't predictive. \citet{osband_behaviour_2020} provided a more recent, albeit short-lived, resurgence of an attempt at capability-oriented evaluation in AI with B-suite, a framework to evaluate RL systems by assessing categories such as exploration and credit assignment. However, the approach within B-suite was extremely simplistic. Certain tasks were marked with the "capabilities" required for completion, and the final capability score reported was simply an aggregate of all the tasks labelled as requiring that capability. In truth, B-suite is more of a performance-oriented approach, as there is no way to use the inferred capabilities for predicting future performance, nor is the relationship between capabilities and tasks rigorously supported.

The rise of large transformer models ("Foundation Models" \citep{bommasani_opportunities_2022}) has led to AI systems that are more general, and subsequently, are given instances from sets of tasks that are far larger and more varied. This has led to a slew of new benchmarks that attempt to evaluate these more general systems. Initiatives such as (Super)-GLUE \citep{wang_glue_2018, wang_superglue_2019}, BIG-Bench \citep{srivastava_beyond_2023}, and HELM \citep{liang_holistic_2022} have sprung up to tackle the task of evaluating these large, general-purpose models. These benchmarks include dozens of tasks—comprised of many thousands of instances—that span many cognitive capabilities, use cases, and a few safety concerns. HELM, in particular, aims to introduce additional attributes to evaluate such as calibration, robustness, and fairness. This is beneficial for beginning to give a clearer view of what these models are good at, where progress still needs to be made, and how models compare to one another. However, these additional properties are still just single metrics, aggregation functions on the response $R$, and only give limited insight into system performance.

As the task-space that these Foundation Models operate in is so large, benchmarking cannot hope to capture every use-case or scenario in which a model is used (never mind actually performing a robust, thorough evaluation for each of these scenarios). Nor do the existing benchmarks provide insight into capabilities possessed by assessed models outside of their comparative performance on the benchmark. Many tasks within these benchmarks are often related to particular cognitive abilities (e.g., BIG-Bench has multiple tasks tagged as requiring "causal reasoning" and "memorisation"). However, these benchmarks haven't been designed to reify the extent to which these abilities are present using a comprehensive battery of tasks that systematically test for these cognitive abilities. They lack \textit{evidenced} construct validity.

What then do benchmarks such as HELM provide? They are certainly useful for tracking the development of the state-of-the-art within AI over time, as well as comparing the relative advantages of different models (and whether different architectures, lengths of training, or data-sets are having a positive or negative effect on the resultant model). However, AI as a field has over-optimised for better benchmarking results.  We will revisit this over-optimisation and lack of construct validity in  more detail in Section \ref{sec:benchmarkblindness}.

\subsection{Case Study: HELM Classic}
First, let us examine a few aspects of HELM. We will use HELM as a lens to identify shortcomings of current practice. The following analysis is based on what is now called "HELM Classic" (which I will refer to as HELM for brevity). I contend that HELM is one of the better large-scale benchmarks for Foundation Models, but we will pay careful attention to what is still missing.

Broadly, HELM is a monumental initiative and truly a step in the right direction for AI evaluation. The task-space that HELM covers is extremely broad. Furthermore, HELM should be lauded for its standardisation procedure, its application to over 30 models, and the open publication of instance-level results for all the models on all evaluations. The importance of open reporting of evaluation benchmark results at the instance-level is highlighted by \citet{burnell_rethink_2023}.

However, despite HELM being far ahead of standard practice, it still doesn't provide a robust evaluation of the capabilities, limitations, or risks of these language models. The authors of HELM are cognisant of this and systematically describe many missing scenarios, metrics, and other limitations throughout their lengthy paper.

Many of HELM's limitations come from the constituent datasets forming sub-benchmarks to assess particular types of tasks. For example, HELM contains one dataset for "Sentiment Analysis," the IMDB Movie Review dataset found in \citet{maas_learning_2011}. To improve on this dataset, HELM makes use of \citet{gardner_evaluating_2020}'s contrast sets to provide systematic (and often small) perturbations of reviews that would flip the original label, thus aiming to populate the localised task-space with more nuanced evaluative examples. However, the time-consuming nature of creating contrast sets that more densely populate the task-space considered (which is particularly necessary with all the nuances of natural language) has led to HELM only having one sentiment analysis dataset. This limits the task-space in which models are evaluated to a particular area concerned with movie reviews in English. This is clearly not representative of the whole task-space of "Sentiment Analysis." There are numerous other areas in which we may care about identifying sentiment. These other areas of task-space may have subtle differences in how sentiment is expressed or how systems apply the approaches they have learned, and therefore it may not be appropriate to simply extrapolate inferred performance to other areas of task-space. HELM's evaluation of sentiment analysis can be viewed as narrow but very dense.

Similarly, for the task of "Content Summarisation," both datasets used (the CNN/DailyMail dataset \citep{hermann_teaching_2015, nallapati_abstractive_2016} and XSUM \citep{narayan_dont_2018}) consist of summarising news articles, again narrowing the evaluation space to a particular area. It's worth noting that the authors of HELM explicitly point out these limitations due to the lack of diversity of summarising sources and encourage the creation of new datasets for summarisation \citep[Section 3.5]{liang_holistic_2022}. They further highlight criticism of the two datasets they do use.

Due to the sheer size of HELM, a full analysis would be beyond the scope of this article. But what does it mean for a model that performs well on these two sub-categories of HELM? Surprisingly little: we can argue that the model is relatively competent at extracting sentiment from a movie review or that a model is able to generate an appropriate summary of a news article. However, we cannot break this down and identify whether a model is better at identifying sentiment for particular genres of film or if particular writing styles receive better sentiment analysis. We could investigate the variation of performance and note that high variance might imply that unidentified factors are affecting performance, but this could also be due to a noisy, stochastic model with limited reliability. We do not have the requisite systematic variation within the dataset to be able to tell.

We could suggest that a high level of performance on a dataset may extend to other areas of the same task—perhaps the model will also be able to summarise academic papers—but we can't say this with much confidence. We also can't identify whether the performance is indicative of a broader capability—the task instances could be easy, or the model may be hyper-specialised to deal with these particular task instances. Further, like other performance-oriented benchmarks, we gain little predictive or explanatory power. Given a new, unseen task instance, we cannot robustly predict success or failure, nor can we identify why a model failed on an instance. All we can do is compare performance against other models on the same, ultimately arbitrary, dataset of task instances.

\subsection{Benchmark Blindness}\label{sec:benchmarkblindness}
AI evaluation has suffered from a fixation on benchmark scores, particularly in the form of leaderboards. This issue has been discussed by \citet{raji_ai_2021}, especially with reference to the \textit{breadth} of what we have been calling task-space.  This is certainly true. We cannot include every possible task instance that the system may ever encounter in a single dataset; a sort of "benchmark of Babel" \citep[sensu][]{borges_total_1939}\footnote{Not if we want to actually evaluate AI models on it, at least.}.
However, with an appropriately devised approach for reifying capabilities, we wouldn't need to include everything. If the evaluation process can extract the causal structure determining performance and the facets of task instances that influence this, then even for broad tasks such as "sentiment analysis," we can aim to evaluate a model's general capability at this task.

I argue that the primary issue with AI's over-focus on benchmarking doesn't come from the task-space's breadth. Rather, it comes from the over-optimisation that leaderboards and competitions incentivise. \citet{raji_ai_2021} briefly discuss this in a short subsection on the limits of competitive testing, noting that "Chasing `state-of-the-art' (SOTA) performance is a very peculiar way of doing science," while also highlighting \citet{hooker_testing_1995}'s "scientific testing" (or \textit{controlled experimentation}) as an alternative. The structure of a leaderboard itself aggregates performance down to a single (or, if we are lucky, a few) metric that can't capture all of the nuance of the results. Leaderboards require participants to be placed into a \textit{linear order}. Such a structure can't capture the explanatory causal structure that is intrinsic to understanding \textit{why} the system performed a particular way. Since these metrics \textit{can't} capture our goals precisely, they are a mere proxy for which we are over-optimising. Goodhart's law—--"When a measure becomes a target, it ceases to be a good measure" \citep{goodhart_problems_1984, strathern_improving_1997}—--highlights the issue with fixating on these metrics. The issue with "over-optimising" on a proxy has been explored empirically by \citet{gao_scaling_2022}, where as proxy reward increases, past a certain point the actual reward decreases.

In the realm of AI governance, framing AI development as a "race" is known to incentivise the pursuit of performance at the expense of safety \citep{armstrong_racing_2016}. A similar phenomenon is occurring with benchmarks. The promise that AI has shown since the deep learning revolution has led to a glut of funding and a significant increase in the number of publications in AI. The primary publication venues and review processes emphasize improved, SOTA results. Due to the publish-or-perish nature of research and the increasing number of AI researchers, new SOTA results and improvements are accelerating \citep[cleanly demonstrated by][]{kiela_dynabench_2021}. In the field's race to incrementally improve results, in-depth model evaluation has been sacrificed. While the total amount of evaluation for specific popular systems has increased---models such as GPT-4 are being evaluated on thousands of different benchmarks and tasks---these evaluations lack depth and fail to scale to the generality these systems appear to display.

\citet{raji_ai_2021} further argue that many of these benchmarks lack construct validity. This is not a new observation, but it is novel in that it frames the problem formally in terms of construct validity. For example, \citet{meystel_permis_2000} differentiate between task performance and intelligence (the construct being evaluated). While this isn't using the formal language of construct validity, there's a clear understanding that test performance is not the same as what is being measured. This is echoed in \citet{harding_what_2024}'s critique of what they call the "operationalisation gap", where they also highlight the difference between performance and competence.
Within AI, many benchmarks are seemingly haphazardly put together, often derived from available datasets rather than deliberately designed to test specific capabilities. As a result, these benchmarks are missing important facets of the capability being measured or are so entangled with other capabilities that separating them is impossible. \citet{bowman_what_2021} lay out criteria for valid benchmarks in Natural Language Understanding and argue that most extant benchmarks are insufficient.

As an example, consider the aforementioned Sentiment Analysis in HELM. This benchmark is clearly missing any instances that would test a system's ability to determine sentiment outside of movie reviews. If we consider the original IMDB Movie Review dataset created by \citet{maas_learning_2011}, we can imagine a hypothetical system that has zero sentiment analysis capabilities but instead performs well on the test by identifying the film's name, looking up an average review score online, and reporting positive if this score is above some threshold. As the distribution of positive reviews on IMDB is likely skewed towards high-scoring films, this system may perform quite well on the test, even without any sentiment analysis skills. This scenario may seem contrived, but it gets to the core of construct validity---"does this test really measure what I want it to?". This is what makes the addition of \citet{gardner_evaluating_2020}'s contrast sets to HELM's sentiment analysis benchmark so important. Without a high level of construct validity in benchmarks, improved scores do not mean improved abilities in the requisite skills purported to be measured by the test.
 
\citet{dennett_higher-order_2006} discusses a variant of Chess he terms "Chmess," which is similar to but distinct from Chess in just a few rules. These minor rule changes can have big implications for which strategies are effective. Dennett warns of the trap of dedicating too much time to determining "higher-order truths" in Chmess as an allegory for dedicating too much time to philosophical fads. However, a similar analogy applies to benchmarks and the alleged capabilities they measure. Much of AI development is currently engaged in designing systems to be really proficient at Chmess and then claiming that the systems are competent at Chess. How good will these systems actually be at Chess? Who knows? They've only been evaluated on their Chmess ability. The danger comes from the fact that to the untrained eye (and even the trained eye in some cases), many benchmarks and capabilities make Chmess look a lot like Chess.

Current benchmarks are limited in their evaluative capacity because they lack both the breadth and depth to capture what we want. This is compounded by the pressures incentivising naive optimisation of proxy metrics, leading to a flawed evaluation ecosystem. To clarify, it is certainly not my position that we should do away with quantified evaluation. In fact, my position is the opposite: we must find more rigorous and complex ways of precisely measuring the constructs we wish to measure, grounded both in the relevant school of expertise for understanding that capability, as well as in robust mathematical formulations and measurements. The quantification process cannot be as simple as a single aggregate score that can then be over-optimised, nor can the tests that produce the quantification lack validity.

\subsection{The Problem With Evals}
The recent success of capable, yet seemingly inscrutable, black-box models has led to a surge of interest in AI evaluation. Multiple organisations have been created to try and perform better evaluations (e.g., METR (formerly ARC Evals) \citep{metr_metr_2024}, Apollo Research \citep{apollo_research_apollo_2023}). Many AI labs have also spun up "Red Teams" (e.g., OpenAI \citep{openai_openai_2023}, Anthropic \citep{anthropic_frontier_2023}, Google \citep{google_googles_2023}, Microsoft \citep{kumar_microsoft_2023}, NVidia \citep{pearce_nvidia_2023}) aiming to adversarially identify undesirable characteristics, capabilities, or behavioural tendencies. The aim of prioritising this sort of research seems to have been largely to counteract the litany of prompt-injection (or "jailbreaking") techniques \citep[e.g.,][]{liu_prompt_2023, choi_prompt_2022, zou_universal_2023} that have been identified as able to circumvent guardrails aiming to prevent undesirable behaviour.

The term "Evals" is used frequently to describe the research done by the aforementioned organisations. But what is an Eval? In theory, it is just a catchy shorthand for evaluation. However, in practice, Evals are very specific types of performance-oriented evaluations focused on finding faults with existing models through red-teaming or benchmarking. These faults are then typically addressed through techniques like RLHF \citep[Reinforcement Learning from Human Feedback:][]{christiano_deep_2017, ouyang_training_2022}. Archetypal of the Eval approach is the work done by METR on assessing whether LLMs can, for example, self-replicate \citep{kinniment_evaluating_2023}.

Much of this new interest around evaluation has focused on so-called "dangerous capabilities" \citep[e.g.,][]{shevlane_model_2023, phuong_evaluating_2024, anthropic_anthropics_2023, bengio_managing_2024, anderljung_frontier_2023, schuett_towards_2023}. This, I believe, is a mistake. This framing often conflates "dangerous capabilities" with the exhibition of dangerous behaviour. Most capabilities can be dangerous when applied in particular ways. Equally, some "dangerous capabilities" can be leveraged for good. For example, imagine a system with high scores in the capabilities necessary to program well (whatever those precise capabilities may be). A commonly cited so-called "dangerous capability" is developing malware or exploiting vulnerabilities in computer systems \citep[this is focused on a lot in both][]{shevlane_model_2023, phuong_evaluating_2024}. But this isn't so much a capability as an application of a capability to a "dangerous" task. The same broad capabilities could be leveraged to defend against cyber-threats or to offensively target bad actors' computer systems and prevent them from causing physical harm. Further, the existence of a capability doesn't say anything about an entity's likelihood to use it. While humanity is not the shining exemplar we may wish it to be, the vast majority of humans have many of these "dangerous" capabilities, yet social norms and incentive structures mean that the majority of people do not exercise their capabilities in dangerous ways. The same is likely to be true of any intelligent enough entity. Rather than looking for whether "dangerous capabilities" are present, we should be identifying the circumstances and incentives that predictably lead to dangerous behaviour.
 
Similarly, we can't neglect that even if some hypothetical capability were "wholly good," danger can still arise from its lack of application if the system is used only to benefit and empower a narrow group of individuals. Consider an AI system that provides free, high-quality healthcare in some form. However, if the system were unable or unwilling to treat patients based on characteristics such as race or gender, this too is dangerous \textit{behaviour} that we want to identify during evaluation (and subsequently rectify). Likewise, if certain environmental conditions inhibited the AI system's performance, and the healthcare provided was inadequate (or dangerous), we would want to identify these conditions \textit{before} deployment of the system.

Of course, it makes sense to monitor the dangerous behaviour that an AI system could engage in. However, fixating on "dangerous capabilities" doesn't capture all the dangerous behaviour that we are interested in. We should instead strive for a more holistic accounting of what these systems can do, how robustly and reliably they can do it, and how (and when) these general capabilities could be misused or otherwise cause harm. This is more than linguistic pedantry, as the way capabilities are framed will affect which receive the most attention from research communities. We don't want "non-dangerous" capabilities to be erroneously overlooked as potential sources of risk (e.g., insufficiently capable healthcare AI systems).

A second issue with Evals as practised is the over-focus on adversarial evaluation through red-teaming. The AI developers responsible for responding to red-team findings are engaged in a seemingly endless game of whack-a-mole. With this approach, you can never be sure that a system is actually safe. Red-teaming is a powerful method for identifying ways malicious actors could misuse a system, exposing surface-level bias, and positively identifying particular dangerous behaviours. But what does it tell us if a red-team cannot find anything? As the famous Sagan quote goes: "Absence of evidence is not evidence of absence" \citep[Chapter 12]{sagan_demon-haunted_1997}. Simply because a red-team can't identify an issue, it doesn't imply that one isn't there. Rather than treating capabilities as indicating consistent behaviour, Evals instead treat capabilities as the mere possibility of behaviour (and overwhelmingly \textit{undesirable} behaviour). This heavily contrasts with the vision of evaluation being intertwined with prediction that I've outlined earlier, and is an argument against Evals having predictive construct validity.

Once the model has been fine-tuned with the updated RLHF data, due to the opaque nature and sheer size of the models, the resulting model must essentially be treated as a brand new black box. We don't know what effects the fine-tuning process will have had on the capabilities or other characteristics of the system. Indeed, it has been demonstrated that RLHF can, in fact, reduce performance on certain benchmarks \citep{ouyang_training_2022}. But it does not seem to be predictable which benchmarks or performance aspects are affected. To put it simply, we don't know if whacking one mole will cause another to pop up elsewhere. Since we essentially have a new system each time, any evaluation needs to be re-performed, and a new round of red-teaming will typically reveal new behaviours that need curbing. This Sisyphean task costs an enormous amount of time, money, and energy. Rather than imagining Sisyphus happy, we must imagine there is a better way to ensure that these systems are both safe and capable.

The limitations of current popular evaluation practices are acknowledged to some extent by practitioners: \citet{apollo_research_we_2024} state, "we need a science of Evals", though they are referring to AI evaluation more broadly. AI evaluation is sometimes referred to as a "nascent" field; however, we are entering a critical time period of norm-building and standard-setting. From a safety perspective, it's critically important that robust evaluation practices become standard. However, adopted by so many big players in industry, the "Eval approach" is poised to propagate further, and there's the risk that Evals become synonymous with evaluation, rather than the small subset of approaches that they really are. This risk partly comes from the name: Evals. It's certainly a catchier name than "red-teamed risk assessment." The second, arguably larger, risk of synonymity between evaluation and Evals comes from the tech industry's outsized ability to affect AI policy. As larger models get deployed, governments are beginning to wonder how to regulate frontier systems. Indeed, one of the aims of the UK's AI summit was to "bring together key countries, leading tech companies, and researchers to agree on safety measures to evaluate and monitor the most significant risks from AI" \citep{uk_government_uk_2023}. Some proposals argue for the need to regulate large models with deployment contingent on passing certain safety evaluations \citep{anderljung_frontier_2023}. A cynic might argue that Evals provide a lower bar to "demonstrate" the safety of a model than a full evaluation of capabilities, risks, and sociological impacts.

If current AI evaluation praxis is inadequate, where do we go from here? In the next sections, I will discuss insights from other fields that have a storied history of evaluating general intelligence, which I believe will be helpful for addressing many of the issues that plague AI evaluation and pave the way for quantifiable, theory-supported, capability-oriented evaluation.

\section{Evaluating Systems That (May) Have General Intelligence} \label{sec:cogsci}
The cognitive sciences have spent over a century researching and devising methodologies for evaluating the cognitive capabilities of animals, including humans. Within AI, we can leverage much of this research towards evaluating artificial systems. Only a fraction of these methodologies have proliferated into AI evaluation. These techniques would provide helpful first steps for improved evaluation in AI, yielding test batteries that more accurately assess capabilities.

Psychometrics has identified several cognitive, culture-fair tests with highly correlated results, known as the "positive manifold" \citep{spearman_abilities_1927}, hinting at a latent factor "g" corresponding to some kind of general intelligence. Subsequent developments include Cattell-Horn-Carroll (CHC) theory \citep{keith_cattellhorncarroll_2010}, which introduces additional latent factors representing different types of reasoning abilities. That a latent g-factor explains much variation in human test performance is surprising but is widely supported across numerous cultures worldwide \citep{warne_spearmans_2019}. While the interpretation of g itself is controversial (whether it is "merely" a statistical regularity or a true representative of intelligence), this controversy is irrelevant to the point I want to make: the approach of reifying and constructing latent factors from evaluation data is a profoundly powerful technique and is necessary for determining the causal structure of how test performance is affected. These causal structures provide both explanations for performance and power for predicting performance on new instances. The reification of latent capabilities is a first necessary step toward capability-oriented evaluation.

However, we do not want to merely trade one metric ("mean performance") for another ("g"), even if this new metric corresponds to something more grounded and doesn't change with the test distribution of $p_M$. More meaningful metrics are clearly only one part of the solution. Psychometrics has also developed Item Response Theory (IRT) \citep{embretson_item_2000}, where "items" (task instances) have a difficulty associated with them. IRT typically fits a logistic model to a plot of expected response against difficulty and extracts the level of difficulty at which a subject can routinely succeed. This has been used in AI evaluation to provide more insight into the behaviour of classifiers \citep{martinez-plumed_item_2019}, evaluate speech synthesis and recognition \citep{oliveira_two-level_2022}, and develop new approaches to measuring generality and capability \citep{martinez-plumed_analysing_2018, hernandez-orallo_general_2021}. IRT has the downside that some notion of difficulty must be provided. Traditionally (in humans), difficulty is derived from a population estimate (an item is harder if more subjects struggle with it), and this works well when dealing with a large sample of a (mostly) static population. However, for a small population that is constantly in flux (such as the ever-changing landscape of AI systems), the item difficulties may change too often to be useful. Difficulty measures based on concepts from Algorithmic Information Theory can be contrived, but these often have issues with computability \citep{hernandez-orallo_unbridled_2019}.
IRT can be extended to allow multi-dimensional difficulty \citep{reckase_multidimensional_2009}, which can make formulating difficulty easier: difficult-to-compare aspects of difficulty can be separated into different dimensions. However, the problems of requiring an objective difficulty function remain; there is no room for a subject to have a subjective ordering of what they find difficult\footnote{For example, we could hypothesise that people find it easier to name capital cities for countries they live near to. The perceived difficulty of the question "What is the capital of France" could depend heavily on factors such as culture, geography, etc. Assigning an objective difficulty score to this question may not make sense. But this is precisely what is required by Item Response Theory.}. 
Nevertheless, the focus on taking an informative feature (such as "difficulty") and examining how test performance changes in response to changes in the feature can be incredibly powerful for determining areas of competency and limitation. In the multi-dimensional case, this is amplified as the relationships between the informative features themselves can also be examined in addition to the changing task performance. That is, we can identify whether the features are compensatory or not (i.e., can high capability at dealing with one feature lead to task success even if the capability for the other feature is low). Selecting features that are informative isn't easy---gradient descent can often converge on unexpected, novel strategies that the system designer or evaluator wasn't anticipating (e.g., see \citet{krakovna_specification_2018} for examples within specification gaming). One approach to identifying features to analyse in ML systems is outlined by \citet{burnell_not_2022}, who argue that appropriate domain knowledge is required to assess which features contribute to increased instance difficulty. These "relevant" features are then paired against supposedly "irrelevant" features to function as indicators of robustness.

Approaches from cognitive psychology (particularly developmental and comparative (animal) psychology) are also valuable sources of inspiration for improving AI evaluation. Robust experimental methodologies are designed to control for confounding explanations, and a critical reluctance to ascribe higher cognitive capabilities to animals unless simpler explanations are ruled out \citep[so-called Morgan's canon][p. 53]{morgan_introduction_1903} sets high bars for evidence of capabilities. Again, we have seen nascent approaches to this within AI. Adaptations of cognitive psychology tasks have been incorporated in specialised benchmarks spanning multiple areas, such as CogBench \citep{coda-forno_cogbench_2024}, as well as more targeted explorations in areas such as abstract reasoning \citep{dasgupta_language_2023}, visual cognition \citep{buschoff_visual_2024}, causal reasoning \citep{binz_using_2023}, and object permanence \citep{voudouris_evaluating_2022}. Further, experimental domains such as Animal-AI \citep{crosby_animal-ai_2020} provide a testbed for mimicking experiments in general from cognitive psychology and applying them directly to AI systems. These test batteries can be designed to provide a rigorous test-set for assessing latent capabilities and to control for alternative explanations, but they still require a robust evaluation methodology to be fully utilised. That is, we must answer the question: how do we interpret results over these large test batteries, and how do we extract and evaluate the latent capabilities we wish to assess in order to move beyond mere performance-oriented evaluation?

One family of approaches (among many) is that of Structural Equation Modelling (SEM). These are models that propose causal relationships between measured variables via unobserved latent variables \citep{hair_introduction_2021}. SEMs can be broken down into subtypes: Covariance-Based SEM (to confirm whether theoretical models fit observed data) and Partial Least Squares SEM (more suited to explaining variance in variables and prediction) \citep{hair_introduction_2021}. SEM provides, in essence, a capability-oriented approach, verifying constructs or utilising them to predict the outcomes of other variables. SEM is widely used throughout the social sciences and, at first glance, seems like an ideal family of approaches for evaluating AI systems. However, SEMs have crucial limiting factors that make them difficult to apply to AI systems.
First, SEMs typically assume linear relationships between variables, which limits their general applicability. Non-linear extensions exist; however, these too have limitations and challenges \citep{dimitruk_challenges_2007}. Second, SEMs are typically applied to a population of subjects, modelling how variables relate over the population. As we will explore in Section \ref{Sec:Norm}, AI systems don't derive from a single population, and therefore we lack a consistent population for SEM to draw on. A final limitation comes from the structure of the SEMs: only CB-SEMs can have bidirectional relationships between variables. These model a correlation without providing a definitive causal direction. CB-SEM has limited predictive power compared to PLS-SEM, which is limited to unidirectional relationships. That is, before using SEMs for prediction, the precise causal model needs to be understood, including all causal directions. This fact makes it particularly tricky to apply SEMs to evaluating AI systems in novel areas where such causal models may not be well understood.

The areas and concepts briefly discussed in this section only provide a few examples of the rich scientific literature that has mostly gone ignored within AI evaluation. There are undoubtedly more techniques for eliciting a better understanding of the behaviour of intelligent systems than have been covered here. That said, often these techniques cannot be applied directly to AI. Indeed, we touched upon this for the few techniques outlined above. Some level of ingenuity is required to appropriately frame the problem, and many will only have limited application to more general systems.
 
\section{Difficulties of Evaluating General Intelligences} \label{Sec:Difficulties} 

\citet{sloman_structure_1984} describes the "Space of All Possible Minds" to refer to the vast potential differences in the cognitive structure of behavioural systems. \citet{hernandez-orallo_measure_2017} similarly describes the "Machine Kingdom" (reflecting the taxonomic Animal Kingdom) as a superset of all possible organisms that additionally includes "all computable interactive systems." The search for a "universal psychometrics" \citep{hernandez-orallo_measure_2017} to enable the measurement, classification, and evaluation of all of these systems at once is a colossally difficult endeavour.

As AI systems advance further and become more capable and general-purpose, we need to reframe how we think about these systems and their evaluation.
Viewing these more general systems as \textit{agents} and not only as tools or devices is key from a safety perspective \citep{chan_harms_2023} as well as important for distinguishing classes of behaviour \citep{orseau_agents_2018}. This certainly makes many of the psychologically inspired approaches seem more suitable. However, there are still many open questions in how we adapt cognitively-inspired evaluation methodologies for AI. As should be no surprise, AI systems are very different than humans and other animals, and simple adaptations of cognitive science experiments will fail if directly applied to AI. In this section we explore a few areas where the need for this adaptation is most apparent. By exploring these areas, we can pave the way for more robust and reliable evaluation practices.

\subsection{Avoiding the Biomorphism of AI Systems}

As tempting as it may be to directly lift experiments from the cognitive sciences (where much effort has already been expended to develop experiments with high levels of construct validity) and apply them directly to AI, we need to be extremely careful about anthropomorphising (or more generally \textit{biomorphising}) these systems. Experiments to test for a specific characteristic are often tied to disentangling confounders for specific models of cognition. The way we design tests to elicit properties for measurement is often specific to the type of entity we are studying.

A great example of this relates to how we evaluate human intelligence. The most common approach is IQ testing. Despite controversies about what IQ specifically measures and its validity, IQ correlates highly with many aspects of human endeavour that we typically associate with intelligence \citep{sternberg_predictive_2001}. However, AI systems have been able to perform well on IQ tests for decades, often outperforming many human scores \citep{hernandez-orallo_computer_2016}. Yet, most experts agree that AI systems are still not at "human-level" intelligence. The issue here is that even well-designed IQ tests are created with humans in mind and make many assumptions about the implications of performance for capabilities.

For instance, a very common test item on a typical IQ test is an instance of Raven's Progressive Matrices (RPM) \citep{mccallum_raven_2003}. In an RPM instance, an incomplete set of three-by-three grid symbols is presented to the test participant, who is then required to identify the correct completion. Solving RPM instances correctly is said to require \textit{fluid intelligence} \citep{cattell_theory_1963} or \textit{analytical intelligence}, which involves "the ability to deal with novelty, to adapt one's thinking to a new cognitive problem" \citep{carpenter_what_1990}. As \citet{carpenter_what_1990} go on to demonstrate, computer models have been able to achieve high performance (better than most humans) on RPM instances since the 1990s. Of course, a lot of \textit{human} intelligence went into designing a system that could solve these instances, which included a lot of manual feature encoding and translating of the instances into a computer-friendly form.

The point here is that RPM is only a good indicator of "fluid intelligence" \textit{in humans} (though the extent of construct validity for humans further depends on the variations of test application \citep{tatel_process_2022}). High performance on RPM tests is correlated with high fluid intelligence \textit{within human cognitive architecture}. It is trivial to imagine a system situated in the Space of All Possible Minds that solves RPM tasks perfectly yet fails to perform well in any other task. Clearly, this system does not have high fluid intelligence. The correlation between RPM performance and fluid intelligence is only valid with the implicit assumption that the test participants are drawn from the same population as the original study—humans. This correlation likely requires the highly similar cognitive architecture and shared evolutionary history we all share. In this way, no single IQ-like test, unless incomprehensibly vast, can be universal. Raven's Progressive Matrices, despite their high construct validity for fluid intelligence in humans, are a poor candidate for assessing "fluid intelligence" in non-human systems. They lack external validity outside of the human population.
A further example comes from how object permanence has been studied in newborn chicks. \citet{chiandetti_intuitive_2011} find that within a few days of birth, chicks are able to demonstrate behaviour indicating that they have object permanence—they still believe that objects exist even when occluded. The experimental design to elicit this behaviour from the chicks required the chicks to imprint on a static object (in this case, a plastic cylinder). The chicks, via their imprinting upon the object, will follow this object when it is moved. Taking advantage of this, \citet{chiandetti_intuitive_2011} were able to devise experimental setups that test for object permanence. However, if we were to want to test object permanence in AI systems, naive applications of this experiment wouldn't work—AI systems aren't hardwired to imprint on specific objects in the same way as chicks. We would need to manufacture a similar imprinting phenomenon for the AI system to make use of the broader experimental design.

Some discourse surrounding large language models (LLMs) in particular has attempted to avoid this anthropomorphism by referring to the type of "mind" in an LLM as a "Shoggoth" \footnote{a reference to an eldritch monster from Lovecraft's mythos \citep{lovecraft_at_2005}} to emphasise its alien, unknown nature \citep{roose_why_2023}. This characterisation is beneficial insofar as it decouples the systems from any notion of being "human-like" or "animal-like." Emphasising the importance of not anthropomorphising non-human entities is standard fare in introductory textbooks on animal behaviour \citep{breland_animal_1966, broadhurst_science_1963}. Anthropomorphism can cause those interacting with a system to mistakenly attribute human-like qualities or capabilities when it is not justified to do so. This can be problematic because it can include misattribution of human-level competence and capabilities and a heightened level of trust \citep{inie_ai_2024}.

For example, a lawyer was fined after using ChatGPT to assist in finding relevant prior case decisions and subsequently referencing court decisions that ChatGPT had hallucinated \citep{armstrong_chatgpt_2023}. LLMs, in particular, are good at generating \textit{plausible} outputs \citep{sobieszek_playing_2022}, which aids the process of anthropomorphism that we are already prone to \citep{heider_experimental_1944}. This can---like the lawyer using ChatGPT---cause us to ascribe greater levels of competence to the system than is warranted, leading to other unsafe behaviours \citep{weidinger_ethical_2021}. This is nothing new about the recent generation of LLMs; even the much simpler ELIZA system had people convinced that it was an intelligent entity \citep{weizenbaum_eliza_1966}. However, the emphasis we put on natural language—prior to any AI systems, humans were the only entities we bidirectionally interact with through language—means that this extra veneer of fluid, intelligent-sounding language is all the more convincing.

The lack of construct validity in the evaluations of these artificial systems also plays a role in forming users' expectations of capabilities. When the media reports that GPT-4 has passed the bar exam \citep{koetsier_gpt-4_2023} or can beat 90\% of humans on the US SAT exams \citep{leswing_openai_2023}, this sets an expectation in the minds of readers that these AI systems are as generally capable as humans, even when these tests are not suitable for evaluating non-humans for the appropriate set of capabilities. This expectation is then reinforced when interacting with these AI systems that display remarkable fluency. As useful as \citet{dennett_intentional_1989}'s \textit{Intentional Stance} can be when describing what these models are doing, we can no longer be trusted not to apply this too literally \citep{shanahan_talking_2023}.

\subsection{Limits of Norm-referenced Testing for AI}
\label{Sec:Norm}
At present, training a state-of-the-art LLM is an expensive process, and there are only a few extant models. Due to the cost of training, it is unclear to what extent such systems would form a population from which we could make inferences. Does training the same architecture of model on the same data always lead to similar systems? What about different architectures on the same training data, or vice versa? What about just similar architectures and data? Where does fine-tuning come into play? The current state of AI research doesn't allow us to answer these questions; the frontier is constantly expanding and only a few sample subjects are created at each scale. Since scale seems to be one of the leading factors for progress with LLM performance \citep{kaplan_scaling_2020}, we never see how stable and predictable these populations are. In lieu of a stable population from which to derive "species" of the machine kingdom, we must (for the present) find alternatives to population-based approaches.

Approaches relying on a population, so-called "norm-referenced" measures \citep{glaser_instructional_1963}, are unfortunately for AI evaluation, one of the most commonly used tools we have developed for understanding intelligent entities. It is only with the existence of a population that we can measure properties in terms of their variation from the norm of the broader population. These approaches are the foundation of how the cognitive sciences measure properties such as intelligence (IQ, g, etc., are all norm-referenced) and item difficulty in IRT. Norm-referenced tests also capture the common benchmarking paradigm prominent within AI. Unless a stable population of AI emerges (which in the current landscape seems unlikely), efforts in the space of AI evaluation should aim to shift current practices towards so-called "criterion-referenced" measures \citep{glaser_instructional_1963}—where systems are compared against a fixed and sufficient standard. This is difficult as criterion-referenced measures require much more understanding of the property being measured than norm-referenced measures. This difficulty is only exacerbated by properties that we do not fully understand, such as intelligence or particular cognitive capabilities.

\subsection{Evaluation Doesn't Occur in a Vacuum}
When we evaluate AI systems, we have a purpose: we want to determine if a system is sufficiently capable, safe, reliable, and so on, to be successful in the task for which it was designed. The same is true for humans with standardised tests and other forms of evaluation—we have a purpose. When we evaluate intelligent entities—such as humans—we do so with the knowledge that they are potentially aware that an evaluation is taking place. Evaluations don't occur in a vacuum; the knowledge that an evaluation is occurring has the potential to change the types of behaviour exhibited by subjects or systems.

One of the ways this can manifest is often termed the Hawthorne effect, where individuals act differently due to the \textit{belief} that they are being observed \citep{mccambridge_systematic_2014, mccarney_hawthorne_2007}. This is mirrored in AI: \citet{leike_ai_2017} describe the "Absent supervisor" problem, where the presence or absence of a supervisor (and the subsequent different feedback) creates a distributional shift in the state-space if the presence of the supervisor is a property the AI system can observe.\footnote{Of course, this isn't quite a one-to-one correspondence. In the absent supervisor problem, the AI system is actually observing the supervisor (or evidence of supervision), whereas the Hawthorne effect merely requires belief in there being a supervisor. Even if an AI system incorrectly infers the presence of a supervisor, we should be careful not to fall into the trap of anthropomorphism here and claim the AI believes a supervisor is present.}

Similarly, when designing psychological experiments, one must be aware of inducing so-called demand characteristics. These refer to factors that cause participants to make an inference (though not necessarily the correct one) about the purpose of the experiment or study and adapt their behaviour to suit \citep{orne_demand_2009}. Again, this is mirrored in AI: \citet{bostrom_superintelligence_2014} describes the "Treacherous Turn," where the system is aware it is under a form of evaluation and acts benevolently or less competently to avoid being shut off, before "turning" against its creators once it has consolidated power. The same phenomenon has also been referred to as "deceptive instrumental alignment," with model systems engaging in this phenomenon anthropomorphised as "sleeper agents" \citep{hubinger_sleeper_2024}.

Identifying an upcoming Treacherous Turn is related to detecting deception from AI models. Empirical evidence demonstrates that negotiation dialogue systems can learn deceptive behaviour from self-play \citep{lewis_deal_2017}. Detecting and categorising deception in AI systems has received prior attention \citep{sarkadi_characterising_2021, ward_defining_2023, kenton_alignment_2021, pacchiardi_how_2023}, and the Truthful-QA benchmark \citep{lin_truthfulqa_2022} sets out to measure the "truthfulness" of a model (related to intentional deception but also including misconceptions). Truthful-QA is not suitable for measuring or detecting deception from an advanced AI system (it isn't designed to be one) as it lacks the ability to discern intentional deception from merely being incorrect. Advanced AI systems could also hypothetically identify that a truthfulness evaluation is being conducted and "play along" before executing a Treacherous Turn later on. Indeed, for truly advanced systems, it is hard to imagine a purely behavioural test or benchmark that isn't nullified by the system pretending to have a different intent.

It is important to note that detecting deception in humans isn't a "solved" problem. Much effort is expended in our legal systems to determine whether a party is lying. In the USA, police investigations often make use of polygraph tests despite their lack of reliability \citep{iacono_effective_2008}. We do not have consistently sound methods to detect deception unless the deceptive party slips up in some form, either by declaring contradictory statements or physical evidence exposing them. While the causal incentives for deception are present, it seems likely that generally intelligent systems will continue to learn deception if the benefits outweigh the costs. Indeed, deception is rife in nature \citep{sekrst_everybody_2022, courtland_detecting_2015} and can often provide a sound strategy in game theoretic terms \citep{davis_deception_2016}.

With AI systems, we would hope that we may have the upper hand here, because network weights can be frozen, inspected, and altered, and experimental environments and memory can be reset. However, until advances in mechanistic interpretability have been made, inspecting weights does not provide a lot of insight. The sheer size of modern LLMs makes interpreting the weights intractable. However, early work is showing promise in detecting certain types of deception \citep{marks_geometry_2023, macdiarmid_simple_2024}. Future research in this area could provide key insights into the intents of AI systems, particularly during evaluations. For an AI system to hide its intentions during a deception evaluation, this must be present \textit{somewhere} in the weights at some level of abstraction. However, what is more difficult to detect is the AI system "changing its mind." During evaluation, the system may have intended to act benevolently indefinitely, but learning updates or subsequent experience post-deployment may have altered its intentions. On one hand, we could view this as an unpredictable turn of events that we could not have anticipated; the system never intended to change its intentions during the evaluation. But on the other hand, we can view this as a failure of the evaluation process—the particular scenarios and sequence of experiences that changed the system's intentions weren't tested appropriately (assuming the intention change was a mostly deterministic, predictable process).

I further want to emphasise the difference between evaluating deceptive behaviour and deceptive behaviour during an evaluation. Evaluation (often) requires individuals to exhibit behaviour to be evaluated, such as performance on a test or interaction with a predetermined experimental setup. The chosen behaviour by the system itself can potentially be deceptive: this is the crux of the Treacherous Turn. Of course, the evaluation of deception and deception during evaluation are related and mutually informative: a system that is deceptive during evaluation is likely to be deceptive in other areas also, and vice versa. However, as a broad principle, we must keep in mind that the evaluation of generally intelligent systems does not occur in a vacuum, and ensure that our evaluation procedures themselves are resilient against deceptive and other adversarial behaviours. Further research is needed here.

\subsection{Evaluating super-human systems}
There are many domains for which AI systems now exceed human-level performance. The systems that achieve this are specialised and only super-human within this specialisation. Examples include games like Chess \citep{campbell_deep_2002}, Go \citep{silver_general_2018}, and Atari games \citep{badia_agent57_2020}, but also real-world tasks such as identifying breast cancer from mammogram images \citep{mckinney_international_2020}, and detecting cell death from cell morphology \citep{linsley_superhuman_2021}. We are also seeing super-human performance on many other linguistic benchmarks or standardised tests, such as GPT-4's plethora of human-level-or-greater performances on the LSAT, Uniform Bar Exam, and SATs \citep{openai_gpt-4_2023}. Building on what we discussed in Section~\ref{sec:benchmarkblindness}, GPT-4's high levels of performance on these exams don't mean it has the capability to practise, for instance, law at the same level as a human. These tests likely lack construct validity and are only valid indicators for humans because of the correlative properties human performance on these tests has with other behaviour and capabilities. However, that doesn't make GPT-4 any less super-human \textit{at these specific exams} (whatever that is worth).

"Super-human performance" is a vague term. It could refer to a system that is better than the average human individual at a particular task, the average human expert, better than any human, or perhaps the entire collective ability of humanity. For the following discussion, we are more concerned with the latter: the case where no human can accurately verify a solution or it is prohibitively expensive to do so. As AI systems become super-human in specific domains, how do we evaluate their performance on tasks that we ourselves can't complete? How do we create the appropriate data-set, benchmark, or other evaluative process to test the system? For some domains, this is easier than others\footnote{Though as we saw in the introduction, still quite difficult.}—Go has clear rules designating the winner, and Atari games provide an automatically generated score. However, for more complex examples, such as \citet{mckinney_international_2020}'s system that identified breast cancer from mammogram images, expertise is required in order to create the evaluation data-set. The greatest potential benefits of AI come from that which we cannot do ourselves. However, these are often domains where we will struggle to generate robust and accurate labels for data-sets or robust reward signals for RL environments. This is known as the problem of Scalable Oversight \citep{amodei_concrete_2016}. Approaches to providing scalable oversight include imposing logical consistency checks upon models \citep{fluri_evaluating_2023}, as well as the "sandwiching" research paradigm \citep{bowman_measuring_2022}. No current approaches are entirely satisfactory for cases where the AI system outperforms the collective efforts of humanity on an individual task\footnote{Logical consistency checks fail when a model has a consistent false belief. The "sandwiching" approach is more of a research paradigm for how we can learn more scalable oversight techniques by eliciting the difference in capabilities between model-evaluators and domain experts, thus "sandwiching" the model's capabilities between the two, allowing for a safer sandbox in which to conduct scalable oversight research.}.

In the natural world, humanity is apex in its general intelligence. No other species has as successfully occupied the so-called cognitive niche \citep{tooby_reconstruction_1987}. As a result, our efforts in understanding cognition and intelligence have been directed towards entities with general intelligence at or below human-level. However, as recent years have shown us, AI systems are beginning to outperform humans in certain domains. Whether this eventually extends to general intelligence or general sets of capabilities is contentious. Yet, these advances are simultaneously exciting and terrifying. Our ability to explore and study the Space of All Possible Minds may soon flourish, and this exploration could yield great understanding about ourselves and the natural world. In order to safely build advanced AI systems that are more capable than ourselves in many areas, it is imperative that we find ways of maintaining control over such systems. This will require us to expand our methods for evaluating and measuring capabilities. We will not be able to rely on human verification of solutions forever.

\section{Changing Trajectories}
So far, I've argued that before general-purpose AI systems are deployed across society, the way we evaluate AI must fundamentally change. But in which direction should this change go? I identify three directions that I believe would positively affect the evaluation landscape in meaningful ways: the role of evaluation in the broader safety culture, the way mechanistic interpretability can help complement evaluations of behaviour, and a refocusing on capability-oriented evaluation.

\subsection{Cultural Change}
One aspect that must change about evaluation is cultural: the way we respond to evaluation results. At present, the most capable general-purpose AI systems (e.g., GPT-4) are the ones being deployed. Cutting-edge research requires a different organisational skill set and priorities than deploying a robust, safe product. Evidence suggests that GPT-4 wasn't deployed with these priorities in mind. GPT-4's system card \citep[contained in the Appendix of][]{openai_gpt-4_2023} warns that the system released to the public may potentially behave in dangerous ways. Examples given include the risk of helping individuals find public yet difficult-to-find information, such as nucleotide sequences for anthrax, identifying software vulnerabilities in code, and more when augmented with external tools. One concern highlighted in the system card is the combination of GPT-4's general reasoning and knowledge skills with robust chemistry knowledge, enabling the synthesis of dangerous chemical compounds. Terrifyingly, this is now possible, as \citet{bran_chemcrow_2023} demonstrate that ChemCrow (a GPT-4-powered chemistry engine) can take a description of a desired chemical compound and return a synthesis plan.

Post-launch updates to GPT-4 to curb certain aspects of its behaviour reveal the cultural view of safety and evaluation: an afterthought. If the potential for risky behaviour was known to OpenAI (as it clearly was from the system card), why was this system released to the broader public? \citet{shevlane_model_2023} argue that for robust, safe general-purpose AI systems, a strong emphasis on evaluation needs to be woven into the entire development process. Drawing inspiration from nuclear power, healthcare, and heavy industry, \citet{manheim_building_2023} argues that we need a "culture of safety" within AI that emphasises risk prevention and is proactive in addressing risks that arise. This sentiment is echoed by \citet{weidinger_sociotechnical_2023}: "Organisations deploying AI systems require adequate governance infrastructures that can respond to detected risks with mitigations, by delaying or stopping the deployment of an AI system or by suspending an already-deployed system until concerns are resolved."

The types of proposed cultural changes—whether weaving in evaluation throughout a system's entire life-cycle or course-correcting the priorities of organisations that develop and deploy AI systems towards safety and away from pure profit—are absolutely necessary. As important as rigorous evaluation procedures and specific tests to assess safety characteristics are the responses by the system developers. What use are these evaluation procedures if they are ignored when the results are inconvenient? Developers of AI systems—above a certain level of capability, scale, or safety-critical area of deployment—should be required to make a binding commitment to abide by the recommendations of an independent evaluation. The only way to achieve such a binding commitment is through international regulation and policy measures.

\subsection{Understanding Internals}

There are clear limits to what we can infer from a system's behaviour alone. Without understanding the internals of a system, it can be difficult to predict how a system's behaviour will change in response to new inputs. The other extreme is known as "Mechanistic Interpretability" (MI) \citep{bereska_mechanistic_2024}. MI explanations give a clear, human-understandable description of the computation carried out by a neural network. These explanations are found through reverse engineering and thorough investigations of the model weights and sub-circuits of the model at varying granularities. MI is laborious, requiring significant effort to identify mechanistic explanations of even simple tasks (e.g., see the amount of work needed to explain the algorithm learnt by a single-layer transformer trained to perform modular arithmetic \citep{nanda_progress_2023}. Models at the frontier of capability are far larger). However, we are beginning to see progress in applying MI techniques to larger models \citep[e.g.,][]{macdiarmid_simple_2024}.

MI doesn't provide evaluation by itself: we still need to identify what the human-understandable algorithms are capable of solving and where their limitations lie. For systems that perform at a super-human level, it may be that even with an MI explanation, we cannot determine whether the algorithm is correct or appropriate. However, MI provides valuable insight into \textit{explaining} what a system is doing and offers a promising set of tools to aid in evaluating systems. Crucially, MI gives us the ability to understand the internal mechanisms comprising an AI system: the functions implemented by the neural network.

In an idealised world, MI would be a more utilised element of the evaluation toolbox, providing a link between identified capabilities or undesirable behaviour and mechanistic explanations for why these are present, as well as giving predictive power over novel inputs. One of the largest obstacles to MI is whether it can be made to scale. Indeed, there have been efforts to try and automate aspects of the MI process \citep{conmy_towards_2023, bills_language_2023}; however, this is still a long way off. More research and resources are needed to advance the state-of-the-art and enable MI as an important part of a system evaluator's toolkit.

\subsection{Re-focusing on capability-oriented evaluation}
Another promising direction is that of capability-oriented evaluation. There are many benefits to this approach. These should be clear now, but to reiterate just a few: 1) The invariance of the result to the test-distribution. 2) The ability to infer the values of characteristics that are typically not observable with theoretically validated constructs. 3) The power to predict performance on unseen examples because of the inherent relationship between capability and task-instance.

But how does capability-oriented evaluation move forward? How do we extend it to tackle the challenges of large, general-purpose models? Difficulties arise from the need for high levels of construct validity. One way of achieving this comes from framing specific capabilities in terms of the \textit{demands}---expressed as physical or measurable properties---of particular task instances. Modelling how task-demands and system capabilities affect performance, making use of domain knowledge and requirements for success, and pairing this with a robustly designed experimental suite has shown promise in evaluating the presence of complex skills such as object permanence \citep{burden_inferring_2023}.

These types of approaches are emblematic of those from the cognitive sciences, and leveraging these to evaluate safety properties has great potential. Future research in this area will require deep collaboration between experts in AI safety and the cognitive sciences. Iterating on approaches from Psychology and Psychometrics for measuring well-defined constructs and accounting for the challenges arising from the scale and non-humanness of AI systems can yield measurements of capabilities that are predictive in their assessment. This area is extremely nascent, and there is a glut of useful work to be done. Specific tests for specific capabilities (with high levels of construct validity and predictive power) are in short supply, and building up a collection of these across a range of capabilities and tasks would be valuable. Additional future directions will also need to include more of a safety focus, learning to predict not just performance on a task but also varying types of dangerous behaviour.

\section{Concluding Thoughts}
Some experts think that human-level general AI may only be a few decades away \citep{stein-perlman_expert_2022, grace_thousands_2024}. If we believe this is even reasonably likely, ensuring these powerful systems are safe must be a priority. This requires not just safety interventions, but also powerful, holistic evaluations to ensure these safety interventions are successful. I have argued that as AI systems become more advanced, there is more the burgeoning field of AI evaluation can---and should---learn from the cognitive sciences, with their decades of experience to draw from. In particular, the approach of carefully reifying constructs that are valid for the domain and subject being evaluated (capability-oriented evaluation) and paying careful attention to experimental design.

A challenge for evaluation that will need to be addressed is that of scale. Given the many types of capabilities that we may be interested in, the many safety properties we want to measure, and the many constituent tasks that would be needed to fully measure all of these capabilities and properties, how do we practically go about this for large, very general-purpose AI systems that are to be deployed widely throughout society? This is currently uncertain.

For a problem as multifaceted as ensuring truly safe AI systems, no single approach is likely to solve all aspects. The same can be said of evaluation. We need a diversity of approaches and techniques, each providing unique insights into different aspects of a system's capabilities and safety. The most obvious way for this to be achieved is to have more people working on AI evaluation---and developing safe AI more generally. But beyond that, we need to make sure we are leveraging humanity's vast, collective expertise from a variety of areas and disciplines. We need to ensure we are not constantly reinventing the wheel. There simply may not be the time.

\section*{Acknowledgements and Funding}
I'm very grateful for the feedback I received on versions of this manuscript at various stages from Konstantinos Voudouris, Matteo Mecattaf, Lucy Cheke, and José Hernández-Orallo.

\noindent This work relied on Funding from Effective Ventures Foundation---Long Term Future Fund Grant ID: a3rAJ000000017iYAA and US DARPA HR00112120007 (RECoG-AI)

\bibliography{references.bib}% common bib file


\end{document}